\documentclass[10pt,twocolumn,letterpaper]{article}
\usepackage[review,algorithms]{wacv}      % To produce the REVIEW version for the algorithms track
\definecolor{wacvblue}{rgb}{0.21,0.49,0.74}
\usepackage[pagebackref,breaklinks,colorlinks,allcolors=wacvblue]{hyperref}
\usepackage{xcolor} % for custom colors
\definecolor{myblue}{HTML}{005faf}

\usepackage{graphicx}
\usepackage{caption}
\usepackage{capt-of}
\usepackage{subcaption}
\usepackage{booktabs}
\usepackage{tabularx} 
\usepackage{makecell} % for line breaks in headers
\usepackage{etoolbox}
\usepackage[english]{babel}
\usepackage[autostyle, english = american]{csquotes}
\MakeOuterQuote{"}

\def\wacvPaperID{903} % *** Enter the WACV Paper ID here
\def\confName{WACV}
\def\confYear{2026}

\usepackage{etoolbox}
\makeatletter
\patchcmd{\@maketitle}
  {\begin{center}}%
  {\vspace{-3.5em}\begin{center}}%
  {}{}

\patchcmd{\@maketitle}
  {\end{center}}%
  {\vspace{-3em}\end{center}}%
  {}{}
\makeatother

\begin{document}


\title{SynchroRaMa : Lip-Synchronized and Emotion-Aware Talking Face Generation via Multi-Modal Emotion Embedding}

%%%%%%%%% AUTHORS - PLEASE UPDATE
\author{Phyo Thet Yee\\
IIT Ropar, India\\
{\tt\small phyo.22csz0009@iitrpr.ac.in}
\and
Dimitrios Kollias\\
Queen Mary University of London, UK\\
{\tt\small d.kollias@qmul.ac.uk}
\and
Sudeepta Mishra\\
IIT Ropar, India\\
{\tt\small sudeepta@iitrpr.ac.in}
\and
Abhinav Dhall\\
Monash University, Australia\\
{\tt\small abhinav.dhall@monash.edu.au}
}

% \begin{figure}
% \centering
%   % \includegraphics[width=\textwidth]
%   \includegraphics[width=15cm]
%     {wacv-2026-author-kit-template/images/figure1.pdf}
%       \caption{We propose SynchroRaMa, an expressive talking face generation framework. Given a single reference image, audio, and a textual description, our model can generate talking face videos featuring lip-synchronized, expressive facial expressions and emotional cues while maintaining identity consistency.  Please check supplementary material for output videos.}
%   \label{fig:synchrorama}
% \end{figure}

\twocolumn[{%
\renewcommand\twocolumn[1][]{#1}%
\maketitle
\begin{center}
    \centering
    \captionsetup{type=figure}
    \vspace{5mm}
    \includegraphics[width=0.9\linewidth]{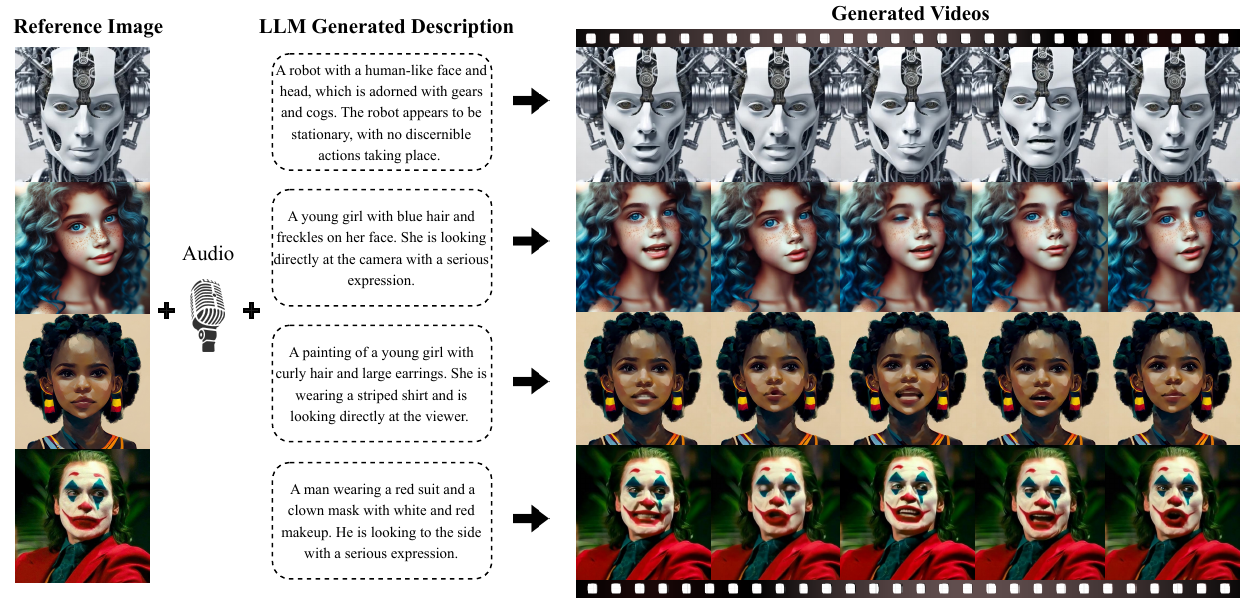}
    \vspace{-2mm}
    % \captionof{figure}{Test caption}
     \caption{We propose SynchroRaMa, an expressive talking face generation framework. Given a single reference image, audio, and a textual description, our model can generate talking face videos featuring lip-synchronized, expressive facial expressions and emotional cues while maintaining identity consistency.}
\label{fig: teaser}
\end{center}%
}]

% \maketitle
% \input{sec/0_abstract}    
% \input{sec/1_intro}
% \input{sec/2_formatting}
% \input{sec/3_finalcopy}

\begin{abstract}
\vspace{-6mm}
Audio-driven talking face generation has received growing interest, particularly for applications requiring expressive and natural human-avatar interaction.
However, most existing emotion-aware methods rely on a single modality (either audio or image) for emotion embedding, limiting their ability to capture nuanced affective cues. 
Additionally, most methods condition on a single reference image, restricting the model’s ability to represent dynamic changes in actions or attributes across time.
%%Furthermore, existing works typically rely on a single reference image to provide visual information. As a result, the model's ability to capture dynamic changes in attributes or actions occurring in subsequent frames is limited.
%Most existing emotion-aware approaches rely on a single modality (image or audio) to provide emotional cues. This limits their ability to capture emotional nuances, as each modality provides only limited information. 
To address these issues, we introduce SynchroRaMa, a novel framework that integrates a multi-modal emotion embedding by combining emotional signals from text (via sentiment analysis) and audio (via speech-based emotion recognition and audio-derived valence-arousal features), enabling the generation of talking face videos with richer and more authentic emotional expressiveness and fidelity.
%This enables the generation of videos with richer and more expressive emotional content. 
To ensure natural head motion and accurate lip synchronization, SynchroRaMa includes an audio-to-motion (A2M) module that generates motion frames aligned with the input audio.
%We also introduce an audio-to-motion module synchronized with the input audio to generate realistic motion frames, ensuring natural head movements and precise lip synchronization. 
Finally, SynchroRaMa incorporates scene descriptions generated by Large Language Model (LLM) as additional textual input, enabling it to capture dynamic actions and high-level semantic attributes. Conditioning the model on both visual and textual cues enhances temporal consistency and visual realism.
%We further incorporate scene descriptions generated by a LLM as additional textual input alongside the reference image. 
%By conditioning the model on both visual and textual cues, our approach generates videos with more detailed actions and attributes. 
Quantitative and qualitative experiments on benchmark datasets demonstrate that SynchroRaMa outperforms the state-of-the-art, achieving improvements in image quality, expression preservation, and motion realism. A user study further confirms that SynchroRaMa achieves higher subjective ratings than competing methods in overall naturalness, motion diversity, and video smoothness. Our project page is available at 
\href{https://novicemm.github.io/synchrorama/index.html}{\textcolor{myblue}{https://novicemm.github.io/synchrorama}}.
% Code and model weights will be released upon acceptance. 

%Experiments demonstrate that SynchroRaMa outperforms state-of-the-art (SOTA) approaches, achieving improvements of 7.6\% in image quality (PSNR), 6.1\% in expression preservation (E-FID), and 4.3\% in motion realism (FVD). Code and model weights will be released upon acceptance.
\end{abstract}

%=============================================================================================================
\vspace{-0.5cm}
\section{Introduction}
\label{sec:intro}

Talking face generation \cite{wei2024aniportrait, ma2024follow, drobyshev2022megaportraits, drobyshev2024emoportraits, xu2024hallo, chen2024echomimic} aims to animate a portrait image by integrating audio. It has gained popularity in various domains such as video games, film industries, social media, digital marketing and education sectors. Existing approaches to talking face generation primarily utilize either GANs \cite{goodfellow2020generative} or diffusion models \cite{dhariwal2021diffusion, ho2020denoising}. GAN-based methods \cite{kr2019towards, prajwal2020lip} use a combination of audio and visual encoders to extract features from speech and video frames, which are then processed by a generator network to produce synchronized lip movements. 
% However, these networks often exhibit limited visual quality, lacking fine facial details and introducing artifacts. 
In contrast, diffusion-based models generate results through an iterative refinement process, leading to higher-quality and more temporally coherent outputs. Despite these improvements, generating a realistic talking face remains challenging, as it requires precise lip synchronization, and head poses with the given speech. In addition to lip synchronization, maintaining the visual coherence and capturing the richness of expression and emotion remain open challenges in talking face generation.

To address these limitations, we introduce SynchroRaMa, a novel framework designed to generate high-quality, emotionally expressive, and lip-synchronized talking faces from the audio input, textual description and a reference image. Most previous works on emotion-aware talking face generation primarily focus on a single modality, such as text, audio or visual cues for emotion embedding. However, relying on a single modality limits model performance, as each modality contains its specific constraints, and individually, none can fulfill all requirements perfectly. Therefore, our work leverages multi-modal emotion embedding by combining textual sentiment analysis, speech-based emotion recognition, and valence-arousal (VA) emotion embedding derived from audio signals. 

Diffusion-based talking face generation approaches typically use a reference network and denoising UNet as their backbone. Visual appearance information is fed into Reference network, which is then integrates with the Denoising UNet during the denoising process. However, providing only visual information is insufficient, as existing methods mostly use a single frame extracted from the input video as the reference image during training. Relying on a single frame to represent the entire video may fail to capture potential changes in scenes, subject's actions, and attributes in subsequent frames. As a result, the generated video may lack detailed appearance consistency. To overcome this, we incorporate textual descriptions that contain changes in scenes (temporal) information as additional input. Recent advancements in visual large language models such as VideoLLaMA2 \cite{cheng2024videollama}, can generate comprehensive textual descriptions of the entire video, capturing changes in scenes, actions and attributes. This textual information complements the detailed visual cues of the extracted reference image, enhancing the visual quality of the generated video. 

Another important aspect in talking face generation is motion consistency. Generated video should exhibit realistic head movement and lip should be synchronized with the speech. To ensure this, we also introduce an audio-to-motion module, producing motion frames driven by the audio. Training the model with these audio-driven motion frames guarantees realistic head movements and accurate lip synchronization. 
Furthermore, we introduce several loss functions in our work, including syncloss, emo loss, facial action units (AU) loss, and attr-action loss. 
% Detailed descriptions of these loss functions are provided in Sec. 3.8.

The \textbf{main contributions} of our paper can be summarized as: 1) We propose a novel talking face generation framework, which leverages multi-modal information, including visual, textual and audio data. 2) We introduce a multi-modal emotion embedding module to enrich emotional expressiveness in the generated videos. 3) We embed LLM generated scene description to make the generation better aligned to context.  4) We present an Audio-to-Motion (A2M) module, designed to generate realistic motion frames synchronized with the audio. Our quantitative and qualitative experiments demonstrate that our approach achieves superior performance compared to the state-of-the-art (SOTA) methods.

%=============================================================================================================

\section{Related Work}

\subsection{Diffusion-based Talking Face Generation}
Diffusion models have recently gained significant attention in talking face generation.
% , producing high-quality animations with fine-grained motion dynamics through iterative noise refinement. 
Hallo \cite{xu2024hallo} leverages the Stable Diffusion Model \cite{rombach2022high} with the ReferenceNet to maintain appearance consistency and introduces a hierarchical audio-visual cross-attention mechanism to align audio and visual features. 
VASA-1 \cite{xu2024vasa} operates in a disentangled latent space to enable precise and expressive facial animations. AniTalker \cite{liu2024anitalker} use universal motion representation to capture a wide range of facial dynamics including subtle expressions and head movements. X-Portrait \cite{xie2024x} employs a conditional diffusion model enhanced with a hierarchical patch-based local control module for accurate and coherent motion transfer. Diff2Lip \cite{mukhopadhyay2024diff2lip} and DiffTalk \cite{shen2023difftalk} use Latent Diffusion Models (LDMs) conditioned on audio features, reference images, masked ground-truth images, and facial landmarks. GAIA \cite{he2024gaiazeroshottalkingavatar} disentangles motion and appearance to preserve identity while enabling speech-driven motion synthesis. 
% The diffusion model is then trained to generate natural motion sequences conditioned on input speech and a reference portrait. 
EchoMimic \cite{chen2024echomimic} and AniPortrait \cite{wei2024aniportrait} employ concurrent training, conditioning simultaneously on audio signals and facial landmarks to generate realistic portrait animations. 
VividTalk \cite{sun2023vividtalk} leverages a 3D hybrid prior to decompose facial expressions and head movements, using a learnable codebook for natural motion and a dual-branch Motion VAE to generate dense motion fields. 
EMO \cite{tian2024emo} utilizes Audio2Video diffusion model, integrating weak condition constraints such as face locators and motion guidance, bypassing the need for intermediate 3D models. 
% Diffused Heads \cite{diffusedhead} uses a single identity image to maintain consistent facial features while incorporating motion frames for temporal coherence, with audio embeddings guiding lip synchronization and expression matching. 
MODA \cite{liu2023moda} employs a mapping-once network with dual attention mechanisms to convert audio into motion representations, where one attention captures accurate lip-sync and the other models natural head and eye movements.

%%Despite the advancements made by previous works, several challenges remain. Many approaches primarily rely on the reference image for appearance consistency, limiting their ability to adapt to varying poses and expressions. Additionally, emotional expression is often overlooked, as most methods condition on audio features without explicitly extracting and preserving emotions. %To address these gaps, we introduce a novel framework that integrates textual prompts describing attributes and actions, capturing richer visual-textual information and ensuring improved appearance in the generated video. Furthermore, we extract emotions from the audio across multiple perspectives, ensuring a more accurate and expressive animation.

%==========================================================================================================

\begin{figure*} [h]
    \centering  \includegraphics[width=\linewidth]{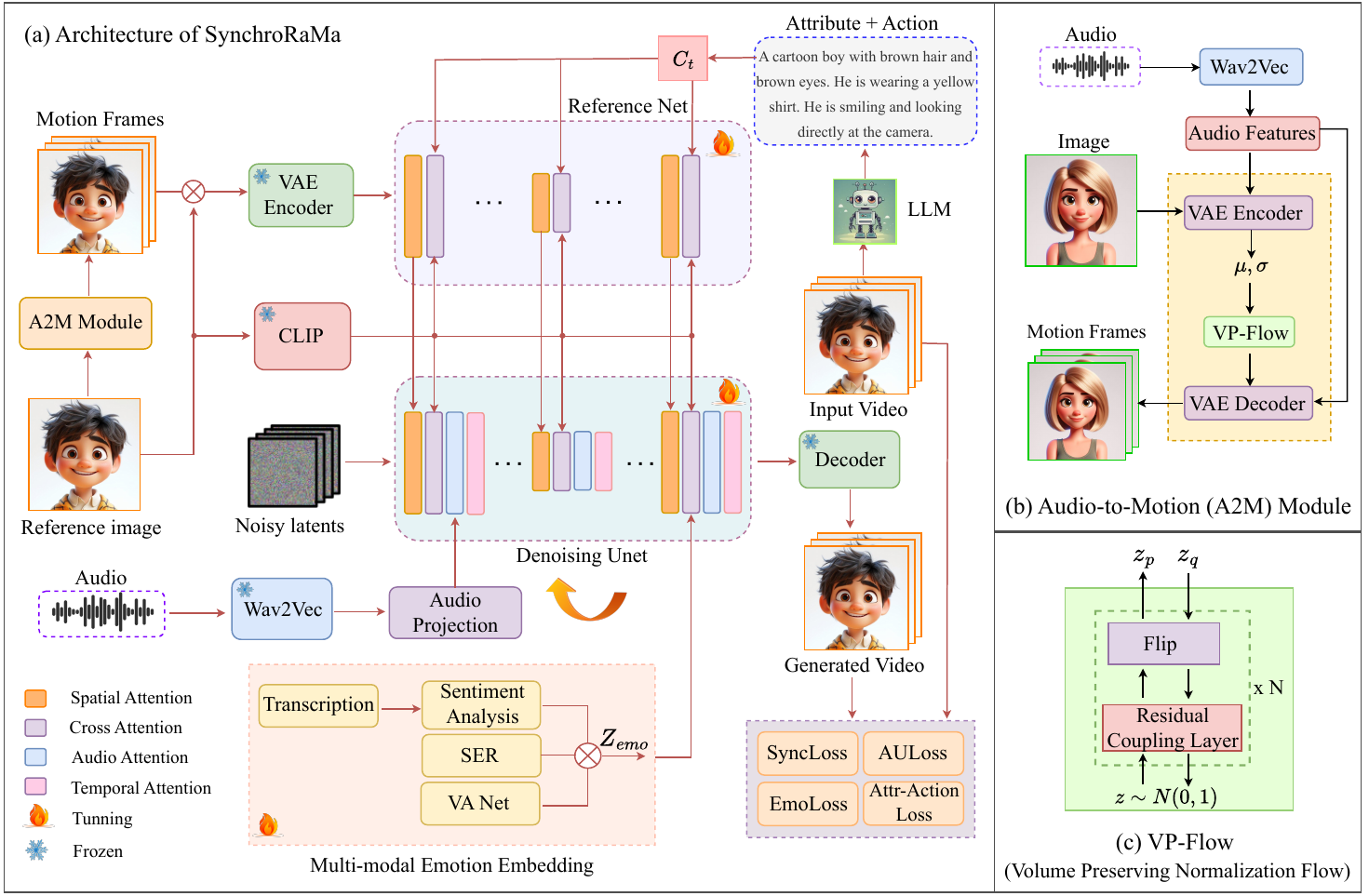}
    % \centering  \includegraphics[width=\linewidth]{wacv-2026-author-kit-template/images/synchrorama_pipeline.pdf}
    %\includegraphics[width=0.8\textwidth]{images/fig2.pdf}
    % \vspace{-0.5cm}
   \caption{(a) Overall architecture of SynchroRaMa. The framework takes a reference image, LLM-generated textual description, and driving audio as input. It operates in two stages. In the first stage, ReferenceNet encodes visual-textual appearance features from the reference image and textual description. Motion frames are also added to ensure temporal smoothness and coherence across frames. In the second stage, these features are integrated with audio projection and a multi-modal emotion embedding within the Denoising UNet during the diffusion process. Motion frames are generated by (b). $C_t$ means CLIP Text Encoder. 
   (b) Architecture of the Audio-to-Motion (A2M) Module. Audio features are extracted using Wav2Vec 2.0, then encoded by a VP-Flow-based VAE, which consists of residual coupling layers and flip operations. (c) Architecture of VP-Flow.}
    \label{fig:pipeline}
    \vspace{-0.5cm}
\end{figure*}

\section{Method}
\subsection{Model Architecture}

Given a reference image, textual description, and input audio, our model generates a lip-synchronized talking face video while preserving natural head movements, facial expressions, emotions, and overall appearance consistency. As shown in Figure \ref{fig:pipeline}, our framework integrates several critical components: Denoising UNet \cite{rombach2022high} as the backbone network, ReferenceNet \cite{hu2024animate} to encode the reference image, proposed Audio-to-Motion (A2M) module to generate the motion frames synchronized with the input audio, and proposed multi-modal emotion embedding module to align the generated video with the emotional content of the audio. The careful incorporation of these components ensures the generation of realistic and contextually coherent talking face videos. The following sections present a detailed explanation of each component.

% \vspace{-3mm}
\subsection{Denoising UNet}

In our framework, we adopt the Denoising UNet architecture from the Stable Diffusion \cite{rombach2022high} 1.5 as the backbone network by incorporating 4 distinct attention mechanisms within each Transformer block (Figure \ref{fig:pipeline}(a)). Specifically, Spatial Attention layer focuses on the important spatial regions, ensuring the model captures the relevant facial features such as mouth, eyes, and expressions while preserving the identity consistency. Audio Attention layer integrates the audio features into the generation process. Cross Attention layer bridges the multiple modalities such as reference image, audio, and textual description by dynamically aligning them throughout the denoising process. Finally, Temporal Attention layer ensures smooth and coherent motion transitions between the consecutive video frames. These integrated attention mechanisms enhance the model's ability to capture and integrate spatial and temporal relationships effectively within the generation process.

\subsection{ReferenceNet}

ReferenceNet employs the same architecture as the Denoising UNet and guides the generation process by embedding both reference image and the textual input (Figure \ref{fig:pipeline}(a)). It ensures the preservation of facial identity and background consistency. Within the ReferenceNet, each Transformer block employs a Spatial Attention mechanism to extract features from the reference image. These features are then passed to the Denoising UNet via the corresponding Spatial Attention layers. Additionally, Cross Attention layer integrates the textual description that contains attributes and actions of the corresponding video as additional conditioning input. This provides semantic guidance to further enhance appearance consistency. This integration of visual and textual conditioning ensures identity and background consistency, as well as high-quality video generation.

\vspace{-1mm}
\subsection{Video Description}

Previous works use one extracted frame from the original video as a reference image during training to provide appearance information (e.g., identity and background details). However, relying on a single image may fail to capture dynamic changes in actions and attributes that occur over time, e.g., a person sitting in the first frame might change position in subsequent frames - a single image cannot reflect these changes. To address this limitation, we incorporate a textual description that represents the entire video alongside the reference image. Specifically, we employ VideoLLaMA2 \cite{cheng2024videollama} to generate a video description that captures fine-grained details such as attributes (e.g., "wearing earrings") and actions (e.g., "turning head"). This textual description is encoded using a CLIP Text Encoder and is incorporated into the model through cross-attention layers in the ReferenceNet and the Denoising UNet (Figure \ref{fig:pipeline}(a)). During training, ReferenceNet processes the reference image and textual description to extract visual-textual appearance features, which are integrated into the Denoising UNet, then fused with emotion and audio features. This conditioning strategy ensures that high-level semantic cues from text influence both attributes and actions throughout the video.

\vspace{-1mm}
\subsection{Audio-to-Motion (A2M) Module}   %%%%%%  UPDATED  %%%%%% 
The main challenge in audio-driven motion generation is to generate temporally consistent and expressive motion sequences that align with the input audio. To address this, we propose a VAE-based Audio-to-Motion module, which takes both a reference image and audio features as input and generates motion frames (Figure \ref{fig:pipeline}(b)). We first extract audio features using wav2vec 2.0, and use them as conditional inputs to the VAE encoder. 
In VAE, the encoder typically outputs parameters $(\mu, \sigma)$ that define a Gaussian distribution, from which a latent variable $z_q$ is sampled as $z_q \sim \mathcal{N}(0,1)$. However, this Gaussian prior tends to force the posterior toward the mean, which limits output diversity and generative power. To overcome this, we follow \cite{ye2023geneface, ye2023geneface++, ren2021portaspeech} and introduce a Volume-Preserving Normalizing Flow (VP-Flow) that transforms the simple latent variable $z_q$ into a more expressive latent representation $z_p$ (i.e, $z_p = f_{VP-Flow}(z_q)$) (Figure \ref{fig:pipeline}(c)). The VP-Flow is composed of a stack of residual affine coupling layers and channel-wise flip operations, which ensure invertibility and maintain the volume of the latent space during transformation. We then pass the transformed latent variable $z_p$, together with the audio features to the VAE decoder, which then synthesizes a sequence of visual motion frames that are aligned with the input audio. Therefore, VP-Flow-based conditioning in our A2M module enhances the diversity and realism of generated motion, while the audio features ensure accurate lip sync and smooth transitions across frames.

\vspace{-0.5mm}
\subsection{Multi-modal Emotion Embedding Module}

Emotion embedding is an important aspect in talking face generation to provide more realistic and expressive facial movements aligned with the driven audio. To achieve this, we introduce a multi-modal emotion embedding mechanism that integrates emotional cues from three modalities - text, audio, and valence-arousal (VA) features (Figure \ref{fig:pipeline}(a)), providing a richer perspective than any single modality alone. First, we transcribe the input audio using Whisper \cite{radford2023robust} and perform sentiment analysis on the resulting text using emotion-english-distilroberta \cite{hartmann2022emotionenglish}. 
% Because we believe that text-based sentiment/emotion analysis could provide additional useful supervision. 
Second, we perform speech emotion recognition (SER) on the audio using \cite{radford2022whisper}. Third, we extract valence and arousal features from the audio using a fine-tuned wav2vec 2.0 \cite{wagner2023dawn}. To ensure accurate extraction of valence-arousal features, we remove background music and divide the audio into $50\%$ overlapping segments to capture temporal variations. The features from each segment are then concatenated to form the final representation. 
% Finally, these three components, textual sentiment, speech emotion recognition, and valence–arousal features, are combined into a single emotion embedding, which is then fed into the denoising UNet through a cross-attention layer. 
Let $E_t$,  $E_{ser}$, and $E_{va}$ denote emotion embeddings from textual sentiment analysis, SER, and VA features, respectively, computed as $E_t = f_{\text{Transcription}}(a)$, $E_{ser} = f_{\text{SER}}(a)$, $E_{va} = f_{\text{VA}}(a)$.
% \begin{align*}
% E_t &= f_{\text{Transcription}}(a), \\
% E_{ser} &= f_{\text{SER}}(a), \\
% E_{va} &= f_{\text{VA}}(a).
% \end{align*}
These embeddings are concatenated to form the final emotion embedding: 
\begin{align}
    E = \text{Concat}(E_t, E_{ser}, E_{va})
\end{align}
This combined embedding is fed into the denoising UNet via a cross-attention layer. By employing multiple modalities, our method captures a broader range of emotional nuances, resulting in a more expressive talking face that closely aligns with the speaker’s intended emotion. 

\vspace{-0.5mm}
\subsection{Loss Functions}
Alongside standard diffusion loss, we introduce auxiliary losses to ensure that the generated video maintains temporal coherence, emotional expressiveness, identity consistency, and semantic alignment with the input modalities. 

\noindent\textbf{Sync Loss.}
This loss enforces precise temporal alignment between generated lip movements and input audio, which is critical for perceptual realism in talking face. We evaluate synchronization using \cite{iashin2024synchformer}, which estimates the temporal offset between audio and video, and detects the location of misalignment. The loss compares the global temporal alignment signals, which are more robust to the local jitter and better reflect the human perception. 
% This approach simplifies the synchronization task and enables effective end-to-end training without requiring low-level phoneme alignment. 
It is defined as:
\begin{equation}
\mathcal{L}_{\text{sync}} = \left( \Delta t_p - \Delta t_{gt} \right)^2 + \left( t_p - t_{gt} \right)^2
\end{equation}
\noindent
where: $\Delta t_p$ and $\Delta t_{gt}$ are the predicted and ground truth temporal offset magnitudes between the audio and generated video, and $t_p$ and $t_{gt}$ are corresponding timestamps where misalignment occurs.

\noindent\textbf{Emo Loss.}
This loss enhances emotionally consistent facial expressions by aligning the temporal emotional dynamics of the generated video with those of the ground truth. Specifically, it encourages the output to reflect similar variations in valence and arousal over time. To compute it, we first remove the background music and divide the audio (from both generated and ground truth videos) into 50\% overlapping segments. Employing overlapping segments ensures smoother emotional transitions and reduces sudden shifts or potential artifacts that may occur from non-overlapping segments. For each segment, we extract valence and arousal values using a transformer-based model \cite{wagner2023dawn}, a fine-tuned variant of wav2vec 2.0 that predicts continuous emotional dimensions (valence and arousal) directly from the audio. This yields a sequence of valence and arousal pairs that captures the evolving emotional state throughout the video. The loss provides fine-grained, temporally-aware supervision, enhancing the emotional expressiveness and coherence of the generated facial behavior. The loss is defined as:
\begin{equation}
\mathcal{L}_{\text{emo}} = \frac{1}{K} \sum_{k=1}^{K} \left[ \left( v^{(k)}_p - v^{(k)}_{gt} \right)^2 + \left( a^{(k)}_p - a^{(k)}_{gt} \right)^2 \right]
\end{equation}

\noindent
where: $K$ is the number of audio segments; $v^{(k)}_p$ and $a^{(k)}_p$ \& $v^{(k)}_{gt}$ and $a^{(k)}_{gt}$ are the VA values of the audio of generated video and of ground truth video for segment $k$.

\begin{table*}[t]      
    \vspace{-1mm}
    \centering    
    \renewcommand{\arraystretch}{1.1} 
    \caption{Comparison with SOTA approaches on the HDTF \cite{zhang2021flow} (top) and MEAD \cite{wang2020mead} (bottom) datasets.}
    
     % \begin{tabular*}{\textwidth}
     % {@{\extracolsep{\fill}} 
     % lcccccccccc@ {\hspace{25pt}}}
     \scalebox{0.92}{
     \begin{tabular*}{\textwidth}{@{\extracolsep{\fill}} lcccccccccc}
    \toprule
       Method & PSNR $\uparrow$ & SSIM $\uparrow$ & LPIPS $\downarrow$ &  FID $\downarrow$ & FVD $\downarrow$ & E-FID $\downarrow$ & F1 $\uparrow$ & Sync $\uparrow$ & $CCC_V\uparrow$ & $CCC_A\uparrow$\\ 
    \midrule
     Hallo & 30.63  & 0.71 & 0.18 & 28.91 & 156.32 & 1.38 & 0.68 & 7.58 & 0.53 & 0.56 \\ 
     EchoMimic & 28.90 & 0.51 & 0.44 &  71.33 & 153.84 & 1.32 & 0.63 & 6.20 & 0.51 & 0.49 \\ 
     VExpress & 29.41 & 0.61 & 0.36 & 58.63 & 350.18 & 1.85 & 0.46 & \textbf{8.07} & 0.32 & 0.30\\ 
     Aniportrait & 30.64 & 0.66 & 0.23 & 42.38 & 449.51 & 1.97 & 0.41 & 3.15 & 0.11 & 0.12 \\ 
    % \hline
    
     Ours & \textbf{32.97} & \textbf{0.73} & \textbf{0.17}  & \textbf{27.67} & \textbf{149.67} & \textbf{1.24} &  \textbf{0.71} & 7.03 & \textbf{0.56} & \textbf{0.58}\\ 
    
    \specialrule{1pt}{0pt}{0pt}

     Hallo & 31.69  & \textbf{0.86} & 0.11 & 31.30 & \textbf{147.27} & 1.41 & 0.65 & 6.25 & 0.54 &  0.51 \\ 
     EchoMimic & 29.07 & 0.68 & 0.42 & 63.25 & 164.75 & 1.50 & 0.61 & 5.98 & 0.53 & 0.50 \\ 
     VExpress & 29.34 & 0.65 & 0.39 & 61.48 & 443.96 & 2.31 & 0.42 & \textbf{7.25} & 0.34 & 0.25 \\ 
     Aniportrait & 30.36 & 0.80 & 0.16 & 51.40 & 508.36 & 2.87 & 0.31 & 2.06 & 0.13 & 0.11 \\ 
     Ours & \textbf{32.21} & \textbf{0.86} & \textbf{0.09}  & \textbf{28.47} & 147.87 & \textbf{1.36} &  \textbf{0.67} & 6.84 & \textbf{0.57} & \textbf{0.54} \\ 
    
    \bottomrule   
    \end{tabular*}
    }
    \label{tab:quantitaive_results}
     \vspace{-3mm} 
\end{table*}

\noindent\textbf{Facial Action Unit (AU) Loss.}
This loss provides fine-grained supervision to enhance the expressiveness and realism of the generated videos. It enables the model to accurately capture subtle facial muscle movements, such as eyebrow raises or lip stretches, which are often overlooked by global emotion descriptors such as valence-arousal values. This targeted supervision helps improve the semantic accuracy and coherence of facial expressions, ensuring that localized facial actions align with the intended emotion. Furthermore, AU loss promotes the temporal and spatial consistency across video frames. It also complements the multi-modal emotion embedding by refining the local expression details, while the embedding captures the overall emotional tone. 
AU loss is computed as the squared L2 distance between the predicted AUs of the generated video and the ground truth AUs from the original video. Formally, it is defined as:
\begin{equation}
\mathcal{L}_{AU} = \frac{1}{T \cdot N} \sum_{t=1}^{T} \sum_{i=1}^{N} \left( AU_{p,i}^{(t)} - AU_{gt,i}^{(t)} \right)^2
\end{equation}
\noindent
where: $T$ is the number of frames; $N$ is the number of AUs; $AU_{p,i}^{(t)}$ is the $i$-th AU prediction at frame $t$; $AU_{gt,i}^{(t)}$ is $i$-th AU ground truth at frame $t$. \\

\vspace{-4mm}
\noindent\textbf{Attr-Action Loss}
This loss ensures that the generated video preserves the high-level semantic attributes and actions described in the original video, e.g., head movements, or appearance cues. 
%This is done by comparing textual descriptions generated from both the predicted and ground truth videos using VideoLLaMA \cite{cheng2024videollama}. The loss penalizes semantic mismatches by measuring the cosine distance between their textual embeddings.
We generate the textual descriptions for both predicted and ground truth videos using VideoLLaMA2 \cite{cheng2024videollama}, and compute the loss as the cosine distance between their textual embeddings:
\begin{equation}
\mathcal{L}_{\text{attr-action}} = 1 - \cos \left( \mathbf{e}_p, \mathbf{e}_{gt} \right)
\end{equation}
\noindent
where: $\mathbf{e}_p$ and $\mathbf{e}_{gt}$ are the textual embeddings of the predicted and ground truth video, respectively; $\cos(\cdot, \cdot)$ denotes the cosine similarity between the two embedding vectors.

%=============================================================================================================

\section{Experiments}

\textbf{Implementation Details.}
We conduct all experiments on NVIDIA A100 GPUs. The model is trained in two stages, each for 30k steps, with batch size  4 and resolution 512x512. In the first stage, we train the model to encode appearance information using a reference image and its corresponding textual description as inputs to the ReferenceNet. Each training sample consists of a 14-frame video clip, from which one frame is randomly selected as the reference frame and another as the target frame. Additionally, motion frames are added to ensure temporal smoothness and coherence across frames. During this stage, the VAE encoder/decoder and the CLIP image/text encoders are kept frozen, while only the ReferenceNet and the Denoising UNet are optimized. Both networks are initialized with weights from the original Stable Diffusion model. In the second stage, the model is trained on full video sequences with audio injection for audio-visual alignment and emotion embedding to capture emotional cues. In both stages, a learning rate of $1e^{-5}$ is used. During inference, the model takes a reference image, its textual prompt, and driving audio as inputs, and generates a video sequence by animating the reference image in sync with the audio. We use DDIM sampling with 40 steps to generate each output video clip.

\noindent\textbf{Datasets.}
%We train our model on VFHQ \cite{xie2022vfhq}, HDTF \cite{zhang2021flow}, and a selection of in-the-wild scene clips shared by Hallo3 \cite{cui2024hallo3}. To improve the quality of training data, we perform preprocessing steps. All videos are resized to $512\times512$ resolution. To ensure the clarity in lip region, we filter out extreme side-profile videos by detecting facial landmarks using MediaPipe \cite{DBLP:journals/corr/abs-1906-08172}. We also exclude videos that contain more than one speaker to maintain speaker identity consistency. After preprocessing, we obtain approximately $80$ hours of video data, with each clip ranging from $3$ to $20$ seconds. All videos are standardized to a uniform frame rate of $25 fps$ to maintain temporal consistency across samples. The audio is resampled to $16 kHz$ and normalized. Background music and noise are removed when detected, as they may interfere with emotion recognition and audio-to-motion mapping accuracy. For evaluation, we use $100$ videos from each of the HDTF and emotion-aware MEAD datasets.
We train our model using VFHQ \cite{xie2022vfhq}, HDTF \cite{zhang2021flow}, and a selection of in-the-wild scene clips shared by  Hallo3 \cite{cui2024hallo3}. To enhance  training data quality: all videos are resized to  512x512 resolution; we filter out extreme side-profile views by detecting facial landmarks with MediaPipe \cite{DBLP:journals/corr/abs-1906-08172} (ensuring clarity in the lip region); videos containing multiple speakers are excluded (maintaining speaker identity consistency). In the end, we obtain around 80 hours of video data, with individual clips ranging from 3 to 20 seconds in length. All videos are standardized to 25 fps to ensure temporal consistency across samples. Audio tracks are resampled to 16 kHz and normalized. When present, background music and noise are removed, as they can negatively impact emotion recognition and audio-to-motion alignment.  For evaluation, we use $100$ videos from each of  HDTF \cite{zhang2021flow} and emotion-aware MEAD \cite{wang2020mead}. 

\noindent\textbf{Evaluation Metrics.}
We employ several metrics to evaluate the performance of our model. To assess the similarity between the generated and ground truth images, we compute PSNR, SSIM \cite{wang2004image}, and LPIPS  \cite{zhang2018unreasonable}. Additionally, we use FID and FVD  to measure how closely the generated data matches that of the actual data. To evaluate the accuracy of facial expressions, we compute: (i) Expression-FID (E-FID) for AUs (following \cite{tian2024emo, chen2024echomimic} - expression parameters are extracted using face reconstruction model \cite{deng2019accurate}); (ii) F1 score for AUs; (iii) Concordance Correlation Coefficient (CCC) for Valence $(CCC_V)$ and Arousal $(CCC_A)$. %on visual data (since our model modifies only visual features and not audio tone in the generation process).  
Following \cite{liu2023moda}, we adopt the confidence score from SyncNet (Sync) \cite{Prajwal_2020} to measure audio-visual synchronization.  

\begin{figure}[t]
  \centering
  \includegraphics[width=1\linewidth]{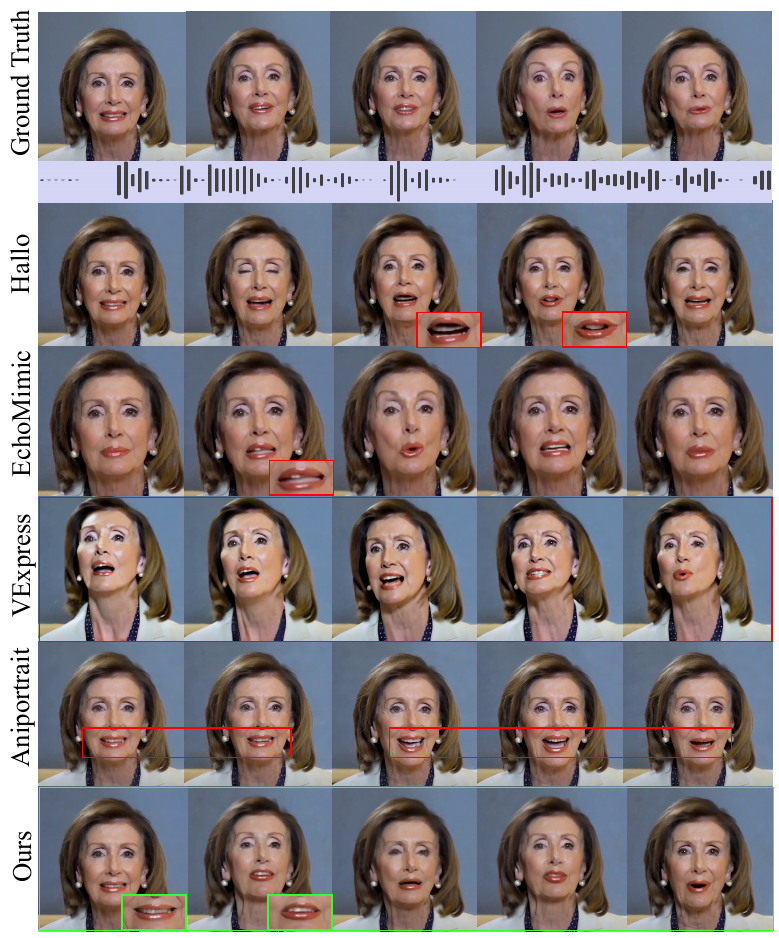}
  \vspace{-6mm}
  \caption{Qualitative comparison with the SOTA approaches on HDTF \cite{zhang2021flow}.  Please zoom in for a better view.}
  \label{fig:qualitative_hdtf}
  \vspace{-5mm}
\end{figure}

%=============================================================================================================

\section{Comparison with state-of-the-art Methods}

\textbf{Quantitative Comparison.}
We perform quantitative comparison with several SOTA, Hallo \cite{xu2024hallo}, Aniportrait \cite{wei2024aniportrait}, Echomimic \cite{chen2024echomimic} and VExpress \cite{wang2024v} on the HDTF and MEAD datasets. We evaluate all methods using their publicly available checkpoints. As shown in Table \ref{tab:quantitaive_results}, our approach outperforms existing methods across all metrics, except for sync confidence on the HDTF and MEAD datasets, and FVD on MEAD. In terms of image-level quality, our method achieves better FID, PSNR, SSIM, and LPIPS scores compared to the SOTA approaches. According to E-FID, F1 scores of AUs, $CCC_V$ and $CCC_A$, our method outperforms competing methods in both expression and emotion preservation. Better performance across these metrics indicates that the generated face videos contain appropriate expressions and emotions, which are reflected in the input audio and reference image. The lower FVD score demonstrates that our approach achieves better video quality. Additionally, our lip-synchronization performance is comparable to that of VExpress and Hallo. 
% Our analysis suggests that the lip synchronization metric is too sensitive to audio, which can sometimes result in unnatural lip movements receiving high scores. Therefore, a higher sync score is not always better result. Therefore higher sync score is not always the best results. 
Notably, despite using less training data than the other methods, our approach achieves better results across various metrics.

\begin{figure}[t]
\vspace{-3mm}
  \centering
  \includegraphics[width=\linewidth]{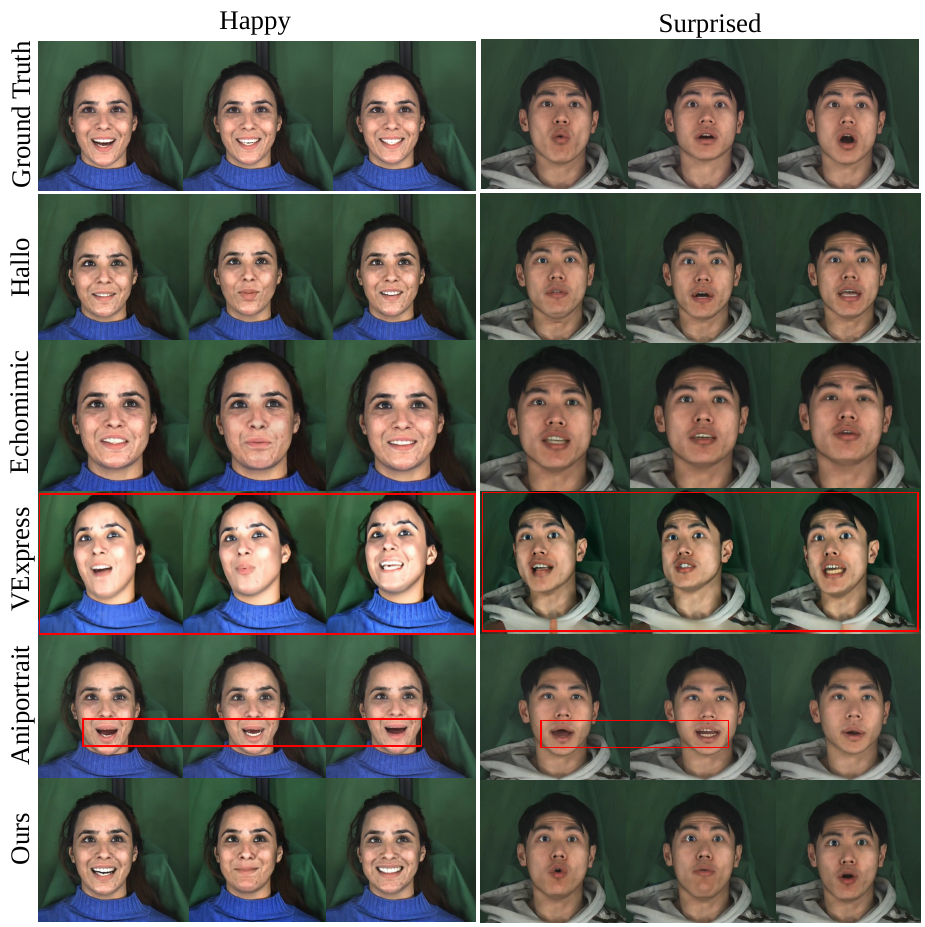}
  \vspace{-5mm}
  \caption{Qualitative comparison with the SOTA approaches on MEAD \cite{wang2020mead}. Please zoom in for a better view.} 
  \label{fig:mead}
  \vspace{-3mm}
\end{figure}

\noindent
\textbf{Qualitative Comparison.}
We perform a qualitative comparison of our proposed method with SOTA approaches \cite{xu2024hallo, chen2024echomimic, wang2024v, wei2024aniportrait} on both the HDTF and MEAD datasets. To evaluate the comparison, we extract frames from the original videos and use the first frame as the reference image. The corresponding audio is also extracted and use it as the driving audio. 
% Figure~\ref{fig:qualitative_hdtf} illustrates the ground truth video alongside the outputs generated by all compared methods, providing a visual comparison of the results.  
Hallo, Echomimic, and VExpress achieve accurate lip sync on both datasets. However, in Figure \ref{fig:qualitative_hdtf}, Hallo produces artifacts in some frames (e.g., a black region around the teeth and unrealistic teeth), while Echomimic exhibits inconsistent motion between frames and unnatural lip (e.g., teeth appearing on the lips). VExpress often fail to generate the correct pose (Figure \ref{fig:qualitative_hdtf}, \ref{fig:mead}), does not maintain identity (Figure \ref{fig:qualitative_hdtf}) and introduces excessive motion. Aniportrait struggles with lip sync accuracy (Figure \ref{fig:qualitative_hdtf}, \ref{fig:mead}), showing minimal or indistinct lip movements across frames. In contrast, our approach generates talking face videos with good overall quality, natural lip movements, and consistent identity. Furthermore, by incorporating emotional awareness in addition to lip synchronization, our model generates videos that are both realistic and emotionally expressive. As shown in Figure \ref{fig:mead}, our method generates emotions that appear more realistic and better aligned with the ground truth than those generated by SOTA approaches.

\begin{table}[h]
    \centering    
    \renewcommand{\arraystretch}{1.1} 
    \caption{Results (in \%) of the user study for comparison of our method with SOTA methods.}
    \vspace{-3mm}
    \scalebox{0.95}{
    \begin{tabularx}{\linewidth}{
        l
        >{\centering\arraybackslash}X
        >{\centering\arraybackslash}X
        >{\centering\arraybackslash}X
        >{\centering\arraybackslash}X
    }
        \toprule
        Method & Lip Sync. & Motion Diversity & Video Smoothness & Overall Naturalness \\ 
        \hline 
        Hallo & 25.73 & 23.61 & 28.53 & 27.35 \\ 
        EchoMimic & 21.11 & 19.02 & 10.82 & 20.30  \\ 
        VExpress & \textbf{26.10} & 7.25 & 5.23 & 7.58 \\ 
        Aniportrait & 2.05 & 5.37 & 2.55 & 2.00 \\ 
        Ours & 25.01 & \textbf{44.75} & \textbf{52.87} & \textbf{42.77} \\ 
        \bottomrule   
    \end{tabularx}
    }
    \label{tab:userstudy}
    \vspace{-3mm}
\end{table}

\noindent\textbf{User Study.}
We conduct a user study to further evaluate the quality of videos generated by our method and by other SOTA methods \cite{xu2024hallo, chen2024echomimic, wang2024v, wei2024aniportrait}. All videos in the study include a balanced representation of genders, with varied ages, poses, and expressions. The study contained 20 participants, all of whom are Master's or PhD students with a background in Computer Science. Each participant is shown videos generated from the same image and audio inputs across all methods. They are then asked to rate each video on a scale from 1 to 5 based on lip sync, motion diversity, video smoothness, and overall naturalness. We collect the ratings and compute the average percentage score for each method. The results are presented in Table \ref{tab:userstudy}. Participants give higher ratings to our approach in terms of video quality, naturalness and motion consistency. Our performance in lip sync is comparable to that of other methods, which is consistent with the findings presented in Table \ref{tab:quantitaive_results}.

% \begin{table} [b]
%     \centering    
%     \renewcommand{\arraystretch}{1.1} 
%     \caption{Results of the user study for comparison of our method with SOTA methods.}
%      \begin{tabular}{ p{1.33cm} p{0.93cm} p{1.3cm} p{1.53cm} p{1.3cm}}   
%     \toprule
%      Method & Lip Sync. & Motion Diversity & Video Smoothness &  Overall Naturalness \\ 
%     \hline 
%     Hallo & 25.73\% & 23.61\% & 28.53\% &  27.35\% \\ 
%     EchoMimic & 21.11\% & 19.02\% & 10.82\% & 20.30\%  \\ 
%     VExpress & \textbf{26.10\%} & 7.25\% & 5.23\% & 7.58\% \\ 
%     Aniportrait & 2.05\% & 5.37\% & 2.55\% &  2.00\% \\ 
%     % \hline
%     Ours & 25.01\% & \textbf{44.75\%}  & \textbf{52.87\%}  &  \textbf{42.77\%} \\ 
%     \bottomrule   
%     \end{tabular}
%     %
%     \label{tab:userstudy}
%      %  
% \end{table}

%=============================================================================================================

\section{Ablation Studies}

We perform the following ablation studies to evaluate contribution of different components of our method:

\begin{figure*}[h]
\vspace{-3mm}
    \centering
    \includegraphics[width=0.85\linewidth]{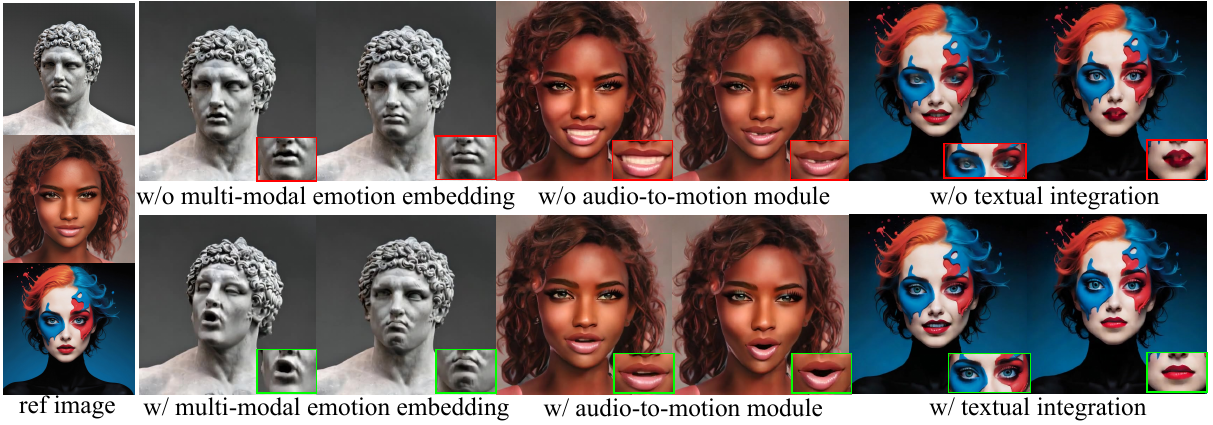} 
    \vspace{-2mm}
    \caption{Effects of Multi-modal Emotion Embedding (left), Audio-to-Motion (A2M) (middle) and Textual Integration (right). } %Note the improved expression and quality of texture in the output.}
    \label{fig:module_ablation}
    \vspace{-3mm}
\end{figure*}

\begin{table}[h]
    \centering
    \vspace{-1mm}
    \caption{Effect of addition of Audio-to-Motion (A2M) Module.}
    \vspace{-3mm}
    \label{tab:ablation_audio-to-motion}
    \scalebox{0.9}{
    \begin{tabular}{ccc}
        \toprule
        Method & Sync $\uparrow$ & FVD $\downarrow$ \\ 
        \midrule
        w/o A2M module & 4.33 & 182.25 \\
        w/ A2M module  & \textbf{6.84} & \textbf{147.87} \\
        \bottomrule
    \end{tabular}
    }
    \vspace{-0.4cm}
\end{table}

\begin{figure}[h]
    \centering
    \includegraphics[width=8.5cm]{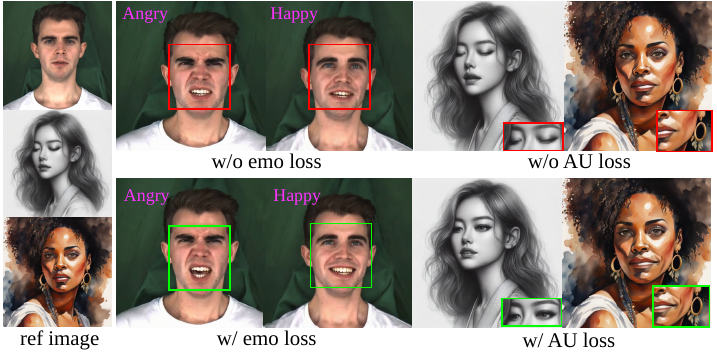} 
    \vspace{-7.5mm}
    \caption{Effects of Emo Loss (left) and AU Loss (right).}
    \label{fig:loss_ablation}
\end{figure}

\begin{table}[h]
    \centering
    \vspace{-0.5mm}
    \caption{Effect of inclduing LLM generated textual description.}
    \vspace{-3mm}
    \label{tab:ablation_LLM}
    \scalebox{0.9}{
    \begin{tabular}{ c c c c }
        \toprule
        Method & PSNR $\uparrow$ & SSIM $\uparrow$ & LPIPS $\downarrow$ \\ 
        \midrule
        \shortstack{w/o textual integration} & 31.30 & 0.62 & 0.15 \\
        \shortstack{w/ textual integration} & \textbf{32.21} & \textbf{0.86} & \textbf{0.09} \\
        \bottomrule
    \end{tabular}
    }
  
\end{table}

\noindent\textbf{Multi-modal Emotion Embedding Module.}
To evaluate the effectiveness of multi-modal emotion embedding module, we perform an experiment (Table \ref{tab:ablation_synchrorama}) by excluding this component. Without the module, the model fails to generate appropriate emotional facial expressions that correspond to audio cues (Figure \ref{fig:module_ablation}, top-left). In contrast, when it is included, the model accurately captures and generates the intended emotional expressions based on the driving audio (Figure \ref{fig:module_ablation}, bottom-left). This is because our multi-modal emotion embedding extracts emotional cues from multiple perspectives, including sentiment analysis, speech emotion recognition, and valence-arousal of the audio.

\noindent\textbf{Audio-to-Motion (A2M) Module.}
When we omit the A2M module and train our model using only the original video frames, the generated video fails to provide correct synchronization and lacks the necessary subject movements (Figure \ref{fig:module_ablation}, top-middle). By adding the motion frames driven by audio and training our model on these audio-aware frames, we ensure accurate lip movements. Based on FVD and Sync scores (Table \ref{tab:ablation_audio-to-motion}), our approach provides better video quality while maintaining the accurate lip synchronization.

\begin{table}[!t]
  \centering    
  \renewcommand{\arraystretch}{1.1} 
  \vspace{-1mm}
  \caption{Effect of adding the multi-modal emotion embedding module and loss functions.}
  \vspace{-2mm}
  \scalebox{0.9}{
  \begin{tabular}{lcccc}
    \toprule
    Method & EFID $\downarrow$ & F1 $\uparrow$ & $CCC_V \uparrow$ & $CCC_A \uparrow$\\
    \midrule
    \shortstack {w/o emo embedding} & 1.43 & 0.58 & 0.48 & 0.50 \\ 
    w/o emo loss  & 1.41 & 0.63 & 0.51 & 0.51  \\ 
    w/o AU loss & 1.38 & 0.63 & 0.53 & 0.48 \\
    Ours & \textbf{1.36} & \textbf{0.67} & \textbf{0.57} & \textbf{0.54} \\
    \bottomrule   
  \end{tabular}
  }
  \label{tab:ablation_synchrorama}
    \vspace{-1em}
\end{table}

% \begin{table} [!t]
%     \centering    
%     \renewcommand{\arraystretch}{1.4} 
%     \caption{Effect of addition of audio-to-motion module.}
%     
%      \begin{tabular}{p{5cm} p{1.2cm} p{1.2cm}}   
%      \toprule
%      Method & Sync. $\uparrow$ & FVD $\downarrow$ \\ 
%     \hline 
%     w/o audio-to-motion module & 4.33 & 182.25 \\ 
%     w/ audio-to-motion module & \textbf{7.10} & \textbf{151.38}  \\ 
%     \bottomrule   
%     \end{tabular}
%     
%     \label{tab:motion_ablation}
% \end{table}

\noindent\textbf{LLM based Textual Integration.}
We evaluate our model without integrating the textual input, relying mainly on appearance cues from the reference image. Without additional semantic guidance from text, the visual quality of the generated video degrades, although lip sync remains unaffected. The model generates inconsistent facial attributes, such as noticeable artifacts in makeup near the eyes and distortions in lip shape (Figure \ref{fig:module_ablation}, top-right). Table \ref{tab:ablation_LLM} shows the quantitative results  with and without textual integration.

% \vspace{-1mm}
\noindent\textbf{Emo Loss.}
Although the multi-modal emotion embedding module enables the model to generate emotional expressions, we found that training the model without emo loss results in less expressive and less accurate facial emotion (Figure \ref{fig:loss_ablation}, top-left). By including the emo loss, which aligns the model’s output with valence and arousal cues derived from audio, the generated expressions become more expressive and realistic (Figure \ref{fig:loss_ablation}, bottom-left). The results in Table \ref{tab:ablation_synchrorama} validate the effectiveness of the emo loss in our model.

% \vspace{-1mm}
\noindent\textbf{AU Loss.}
When we exclude the AU loss, the generated faces exhibit limited facial actions; for example, lip region appears flat and expressionless, and eyes remain closed despite the emotional context (Figure \ref{fig:loss_ablation}, top-right). With AU loss, the model is able to generate fine-grained facial movements, such as the appearance of expression lines near the mouth and properly opened eyes, even when the reference image shows closed eyes (Figure \ref{fig:loss_ablation}, bottom-right). The results in Table \ref{tab:ablation_synchrorama} show that AU loss enhances the realism and expressiveness of the generated videos.  

% \begin{table} [!t]
%     \centering    
%     \renewcommand{\arraystretch}{1.4} 
%     \caption{Effect of integration of LLM generated textual description.}
%     
%      \begin{tabular}{p{3.5cm} p{1.3cm} p{1.2cm} p{1.2cm}}   
%     \toprule
%      Method & PSNR. $\uparrow$ & SSIM $\uparrow$ & LPIPS $\downarrow$  \\[0.2ex]
%     \hline 
%     w/o textual integration & 31.30 & 0.62 & 0.15\\ 

%     w/ textual integration & \textbf{31.53} & \textbf{0.72}  & \textbf{0.13} \\ 
%     \bottomrule   
%     \end{tabular}
%     
%     \label{tab:text}
    
% \end{table}

%=============================================================================================================

\section{Conclusion, Limitations and Future Work}

We propose a novel framework that effectively integrates multi-modal emotional nuances with audio-driven motion modules to generate high-quality, lip-synchronized talking face video. By conditioning the model on visual and textual info, we provide better visual details. Comprehensive experiments, ablation studies \& user evaluations demonstrate our model outperforms SOTA. Results show that SynchroRaMa is an effective tool for creating high-quality, emotionally rich, and lip-synchronized talking face videos.

While our proposed approach achieves promising results, it has some limitations which we will address in future work. First, since our model is trained mainly on portrait images, it is currently unable to generate full-body talking videos. Additionally, because the model was trained only on English language data, its performance on other languages needs to be evaluated.

% Here is my embedded video:

% \includemedia[
%   width=0.6\linewidth,
%   height=0.4\linewidth,
%   activate=onclick,
%   addresource=images/jocker3.mp4,
%   flashvars={
%     source=images/jocker3.mp4
%    &autoPlay=true
%    &loop=true
%   }
% ]{}{VPlayer.swf}
%=============================================================================================================

%=============================================================================================================

% \begin{figure*}
%   \centering
%   \begin{subfigure}{0.68\linewidth}
%     \fbox{\rule{0pt}{2in} \rule{.9\linewidth}{0pt}}
%     \caption{An example of a subfigure.}
%     \label{fig:short-a}
%   \end{subfigure}
%   \hfill
%   \begin{subfigure}{0.28\linewidth}
%     \fbox{\rule{0pt}{2in} \rule{.9\linewidth}{0pt}}
%     \caption{Another example of a subfigure.}
%     \label{fig:short-b}
%   \end{subfigure}
%   \caption{Example of a short caption, which should be centered.}
%   \label{fig:short}
% \end{figure*}

{
    \small
    \bibliographystyle{ieeenat_fullname}
    \bibliography{main}
}

\end{document}